\documentclass[11pt,a4paper]{article}
\usepackage[hyperref]{eacl2021}
\usepackage{times}
\usepackage{latexsym}

\usepackage{microtype}

\aclfinalcopy % Uncomment this line for the final submission
\usepackage{graphicx}
\usepackage{booktabs}
\usepackage{subfig}

\title{From Stock Prediction to Financial Relevance: Repurposing Attention Weights to Assess News Relevance Without Manual Annotations}

\author{Luciano Del Corro\thanks{\hspace*{2mm}equal contribution} \qquad Johannes Hoffart\footnotemark[1]\textbf{}\\
  Max Planck Institute for Informatics, Saarbr\"{u}cken, Germany \\
  \texttt{\{corrogg,jhoffart\}@mpi-inf.mpg.de}}

\date{}

\begin{document}
\maketitle
\begin{abstract}
We present a method to automatically identify financially relevant news using stock price movements and news headlines as input. The method repurposes the attention weights of a neural network initially trained to predict stock prices to assign a relevance score to each headline, eliminating the need for manually labeled training data. Our experiments on the four most relevant US stock indices and 1.5M news headlines show that the method ranks relevant news highly, positively correlated with the accuracy of the initial stock price prediction task.
\end{abstract}

\section{Introduction}
Events such as lawsuits, the unveiling of a newly discovered technology, the introduction of new legislation, or previous market movements can have a significant impact on stock prices. A quick and informed reaction to such an event is crucial for financial analysts.

Information overload is pervasive in the financial industry, hindering the analysts' ability to incorporate the most relevant events into their decision process. One of the main natural language understanding challenges across many industries is to prioritize incoming information, reducing the risk of missing important events.

In this paper, we propose a novel method to identify relevant financial news. The key insight is that this can be achieved without relying on manually created relevance judgments, instead leveraging the correlation of news events and stock prices. The core idea is to train an attention-based neural network on the stock prediction task, using the price movement as label. The input of the network is a set of events in the form of news headlines (embedded using BERT \cite{devlin:19}), and the output is the price movement of a specific stock index with respect to the previous day (i.e., DOWN, STAY, UP), mediated by an attention layer \cite{bahdanau:2015}. The layer acts as an input selector, computing the weight for each headline on a given day. These weights are repurposed to score and thus rank news headlines according to financial relevance. As each weight solely depends on the headline itself, we can use it to compare headlines across the entire dataset.

We evaluated our method on the most prominent US stock indices (S\&P500, Dow Jones, Russell 1000, and Nasdaq) and 1.5 million headlines (1994-2010) from Gigaword \cite{graff2003english}. A first automatic evaluation confirmed a positive correlation between stock prediction accuracy and relevance scores (via the attention weight). In a second, manual evaluation, we labeled 1000 headlines and found that the method ranks relevant events highly: A network trained on the Dow Jones stock index prices, for example, resulted in 89\% relevant events in the top 200 ranks, compared to only 19\% relevant events in a uniform sample of headlines.

\section{Related Work}

{\setlength{\parindent}{0cm}{
\textbf{Stock price prediction from news.} The feasibility of predicting stock prices from news has been debated \cite{merello:2018} as the news can affect the price before it is published. In our case this is not an issue as it only matters that the price change is reflected in the news, not the timing; news about price movements are indeed relevant. Multiple approaches explored alternatives to extract price signals from news such as sentiment analysis, semantic parsing, etc~\cite{Gidofalvi:2001,Schumaker:2009,xie:2013,knosys:2014,peng:2016}. Here we use contextualized embeddings plus attention to extract those signals.
}}

{\setlength{\parindent}{0cm}{
\textbf{Extraction of financial events.} Previous work focused on the explicit representation of events, either in a canonical or semi-canonical way \cite{ding:2014,ding:2015,ding:2016,peng:2016,Jacobs:2018}, or non-canonicalized \cite{Dor:2019,Shi2019}. Events are represented as structured facts \cite{ding:2014}, via embeddings \cite{ding:2015} or keywords \cite{Shi2019}, and are usually pre-selected based on explicit company mentions \cite{Shi2019}. Most approaches differ from ours in that events are input for stock prediction as ultimate goal, while we use stock prediction to identify relevant events. Our end-to-end approach allows us to automatically select the relevant events avoiding involved preprocessing, compromise on the representation of the event, or the use of an underlying extraction system.
}}

{\setlength{\parindent}{0cm}{
\textbf{Attention as explanation.} A debate has erupted around the idea of using the attention mechanism \cite{bahdanau:2015} to explain output. \citet{serrano:2019} and \citet{sarthak:2019} concluded that attention weights should not be used to explain a decision, and \citet{wiegreffe:2019} developed a set of tests to determine weight consistency. In our specific case, we found that the results on the stock price prediction are related to the attention weights performance as relevance scores, and have merit when used for ranking. However, we acknowledge the need to go into deeper analysis to understand score stability in future work.
}}

\section{Training the attention layer for event scoring}

The idea is to make use of attention to learn a headline relevance weight by predicting the stock price movement. Once the network is trained we can use the unnormalized weights of the attention layer as a global relevance score for the news headlines. 

\begin{figure}[h]
    \centering
    \includegraphics[scale=0.68]{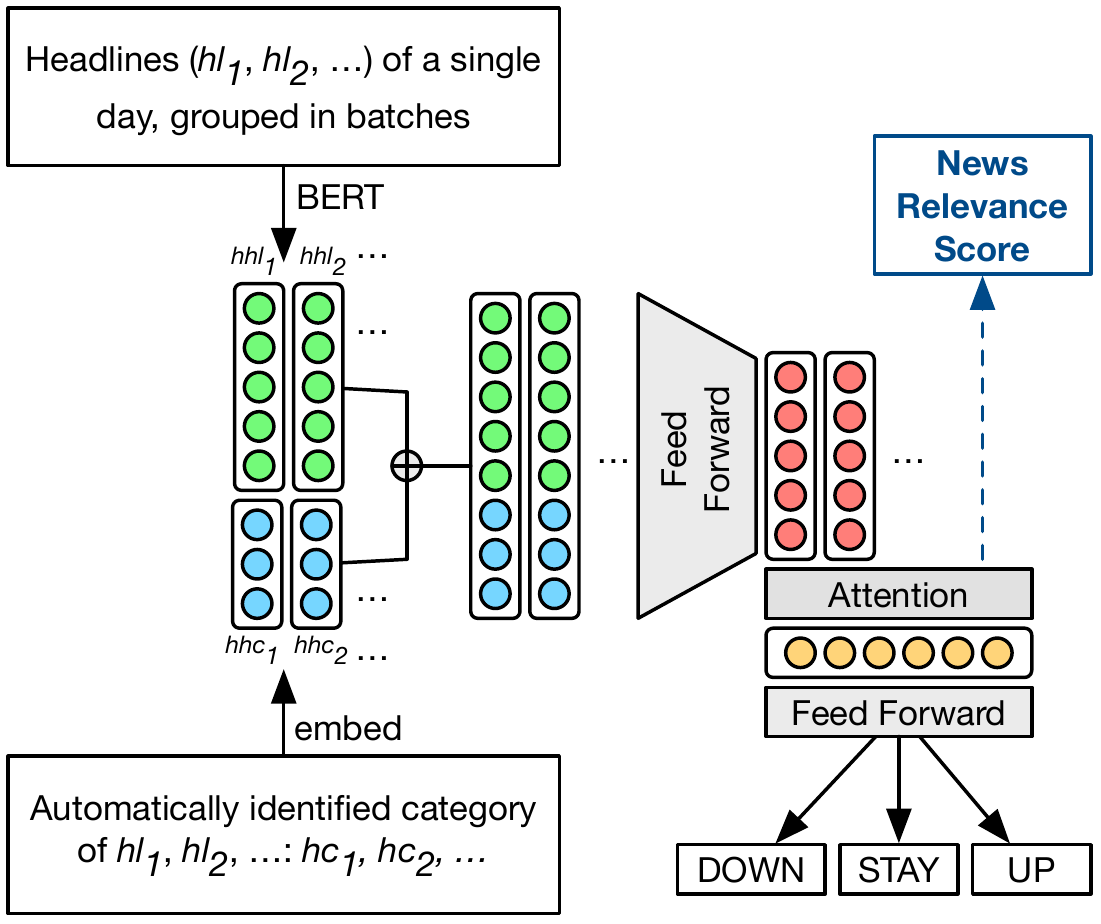}
    \caption{Neural Network Layout}
    \label{fig:network}
\end{figure}

As input we use all daily headlines and their categories. The output is DOWN, STAY, UP with respect to the next trading session open price. The full network is displayed in Figure~\ref{fig:network}.

Each headline $hl_1, \dots,hl_k$ consisting of a (padded) sequence of $N$ tokens $\{w_i\}_{i=1,\dots, N}$, is encoded into vectors ${\{\mathbf{hhl}_i}\}_{i=1,\dots,N}$ of length $768$ via the BERT-base-uncased model pooled output:
\[
\mathbf{hhl}_i = \mathrm{BERT}(hl_i)
\]
Each headline comes with a category label $hc_1, hc_2,\dots$ (details in Section~\ref{par:newsclass}) embedded into a randomly initialized vector of length 30
\[
\mathbf{hhc}_i = \mathrm{embed}(hc_i),
\]
Both vectors are concatenated
\[
\mathbf{h}_i = \mathbf{hhl}_i \oplus \mathbf{hhc}_i
\]
and projected to a vector $\mathbf{h}_{p_i}$ of length 100 by a fully connected feed forward with ELU activation
\[
\mathbf{h}_{p_i} = \mathrm{FF}_{\mathrm{ELU}}(\mathbf{h}_i)
\]
Following \citet{yang:2016}, an attention layer computes normalized weights for each headline in the input day, $\mathbf{H}_{p_d}:=\{\mathbf{h} _{p_i}\}_{i=1,\dots,k}$, and aggregates them according to those weights. 
\[
\mathbf{h}_{a_d} = \mathrm{Attention}(\mathbf{H}_{p_d})
\]
The final label $l_i$ (DOWN, STAY, UP) is computed using a feed forward layer with softmax activation
\[
l_i = \mathrm{FF}_{\mathrm{softmax}}\textbf{}(\mathbf{h}_{a_d})
\]

Every input layer is normalized, and weights are initialized using \emph{He}. The dropout rate is 0.25.  We use the BERT out-of-the-box optimizer\footnote{https://github.com/tensorflow/models/tree/master/official/nlp/bert}. All weights are fine-tuned on the task. We ran on 3 Tesla V100 with a total batch size of 15. The model has a total of $\sim$110M parameters.

\section{Evaluation}

{\setlength{\parindent}{0cm}{
\textbf{Dataset.} AP headlines of English Gigaword~\cite{graff2003english} and the most prominent US stock indices: S\&P500, Dow Jones, Nasdaq, and Russell\footnote{\url{https://finance.yahoo.com/quote/\%5EGSPC/history?p=\%5EGSPC}}, totaling 3777 trading sessions (1994--2010) and 1,532,260 headlines, with a daily average of 405.68, a standard deviation of 134.49, a minimum of 1 and a maximum of 1213. 
}}

{\setlength{\parindent}{0cm}{
\textbf{News Classification.}\label{par:newsclass} We trained a classifier on TagMyNews~\cite{vitale:2012} to classify the headlines into 6 categories: 'business', 'entertainment', 'health', 'sci-tech', 'sport', 'us' and 'world'. The input of the BERT based model is a single headline and the output a class score (dropout=0.25, batch size=120, maximum headline length=15 tokens). The best model F1 was 0.85 (20\% test size), in line with state-of-the-art~\cite{zeng:2018}. We assigned a single class to each headline (0.5 threshold). Table~\ref{news_classification} shows the distribution per category.
}}

\begin{table} [!h]
\begin{center}
 \scalebox{0.8}{
  \begin{tabular}{lcc} \toprule
    \textbf{Category} & \textbf{Number of articles} & \textbf{\%} \\ \cmidrule{1-3}
     world & 596,899 & 38,96\\
     sport & 275,585 & 17.99\\
     business & 231,083 & 15.08\\
     us & 211,570 & 13.81\\
     unclassified & 66,891 & 4.37\\
     entertainment  & 54,607 & 3.56\\
     sci-tech & 54,057 & 3.53\\
     health  & 41,568 & 2.70\\
     \bottomrule 
  \end{tabular}
  }
 \caption{Distribution of news per category}
  \label{news_classification}
\end{center}
\end{table}

{\setlength{\parindent}{0cm}{
\textbf{Preprocessing.}\label{par:preprocessing} Given resource constraints, we are limited to 115 headlines per day, with a maximum length of 15 tokens. To account for more than 115 headlines, we created stratified subsets on headline categories to generate several data points per day. We discarded days with less than 25 headlines for the four most prominent categories, dropping 511 data points (13.53\%), and removed headlines with less than 20 characters. 
}}
To assign price movement labels, we set thresholds that minimize the distance between the majority and minority class to balance the distributions. We searched for a symmetric threshold between [0.1\%, 1\%] at 0.1 intervals, see Table~\ref{stock_class_distribution} for the final values.

\begin{table} [!h]
\begin{center}
 \scalebox{0.8}{
  \begin{tabular}{lcccc} \toprule
    \textbf{Stock Index} & \textbf{Threshold} & \textbf{DOWN} & \textbf{STAY} & \textbf{UP}\\ \cmidrule{1-5}
     S\&P500  & +/- 0.3\% & 30.91\% & 33.61\% & 35.48\%\\
     Dow Jones  & +/- 0.3\% & 30.53\% & 33.23\% & 36.23\%\\
     Russell 1000  & +/- 0.3\% & 29.47\% & 34.38\% & 36.15\%\\
     Nasdaq  & +/- 0.3\% & 30.48\% & 29.83\% & 39.69\%\\
     \bottomrule 
  \end{tabular}
  }
 \caption{Thresholds and class distributions}
  \label{stock_class_distribution}
\end{center}
\end{table}

\subsection{Event relevance}\label{stock-price-prediction}

The goal is to understand if the attention layer's unnormalized weights can be used to generate meaningful global news relevance scores; we understand meaningful as a score that favors news reflecting stronger price movement signals, either ex-ante or ex-post the price change, as in both cases the news would be relevant; we do not control for endogeneity. We ran two experiments: one to understand which news categories provide better signals, and the second to check if they effectively receive higher scores. We also performed a manual evaluation over the top 200 headlines for each index plus a uniform random sample, totaling 1000 headlines.

{\setlength{\parindent}{0cm}{
\textbf{Categories with stronger signals.}  We ran the network on each news category separately to predict the price movement. We selected the model with the maximum accuracy over 20 epochs. Table~\ref{stock_prediction} shows the results for all categories. 'business' headlines are more informative, achieving the highest accuracy on most of the stock indices. Interestingly, 'sci-tech' news is the best category for Nasdaq, which specializes in technology. However, the top accuracy for this index lags well behind the others. This fact will be reflected in the relevance scores in the following experiment.
}}

\begin{table} [!h]
\begin{center}
 \scalebox{0.70}{
  \begin{tabular}{lcccc} \toprule
    \textbf{News Category} & \textbf{S\&P500} & \textbf{Dow Jon.} & \textbf{Russ. 1000} & \textbf{Nasd.} \\ \cmidrule{1-5}
     business  & \textbf{57.88} & \textbf{61.97} & \textbf{55.92} & 43.64\\
     us  & 40.13 & 42.02 & 39.59 & 38.45 \\
     world  & 41.83 & 39.66 & 38.73 & 44.89 \\
     sports  & 38.94 & 36.36 & 38.94 & 44.09\\
     sci-tech  & 36.74 & 37.05 & 36.90 & \textbf{44.96} \\
     entertainment  & 34.57 & 37.98 & 38.14 & 42.33\\
     health  & 34.57 & 34.26 & 35.81 & 40.78\\
     \hline
     all & 52.92 & 54.99 & 54.49 & 45.22 \\
     \bottomrule 
  \end{tabular}
  }
 \caption{Max stock prediction accuracy per category}
  \label{stock_prediction}
\end{center}
\end{table}
\vspace{-0.4cm}

% \begin{figure*}[!t]
%   \centering
%   \subfloat[Global\label{fig:global}]{%
%     \includegraphics[scale=0.53]{gfx/Global_True_top20000_news_ranking}
%   }\\
%   \subfloat[S\&P500\label{fig:sp500}]{%
%     \includegraphics[scale=0.53]{gfx/S_P500_True_top20000_news_ranking}
%   }
%   \hfill
%   \subfloat[Dow Jones\label{fig:dowjones}]{%
%     \includegraphics[scale=0.53]{gfx/DOWJONES_True_top20000_news_ranking}
%   }\\
%   \subfloat[NASDAQ\label{fig:nasdaq}]{%
%     \includegraphics[scale=0.53]{gfx/NASDAQ_True_top20000_news_ranking}
%   }
%   \hfill
%   \subfloat[Russell 1000\label{fig:rui}]{%
%     \includegraphics[scale=0.53]{gfx/RUI_True_top20000_news_ranking}
%   }\\
%   \caption{Event Detection Results}
%   \label{fig:news-ranking}
% \end{figure*} 

\begin{table*} [!h]
\begin{center}
 \scalebox{0.70}{
  \begin{tabular}{lcccc} \toprule
    \textbf{Stock Index} & \textbf{@Rank 10} & \textbf{@Rank 100} & \textbf{@Rank 1000} & \textbf{@Rank 2500}\\
    \cmidrule{1-5}
     S\&P500 & 90.00\% / +471.09\% & 91.82\% / +482.62\% & 87.50\% / +455.22\% & 81.30\% / +415.88\% \\
     Dow Jones & 85.00\% / +439.36\% & 63.00\% / +299.76\% & 40.95\% / +159.84\% & 29.02\% / +84.14\% \\
     Russell 100 & 100.00\% / +534.54\% & 92.33\% / +485.90\% & 87.40\% / +454.59\% & 81.84\% / +419.31\% \\
     Nasdaq & 0.06\% / -61.93\% & 14.20\% / -9.90\% & 16.46\% / +4.45\%  & 16.512\% / +4.78\% \\
     \bottomrule 
  \end{tabular}
  }
 \caption{Percentage of 'business' news at rank / Percentage increase compared to base distribution of 15.76\%}
  \label{tab:news-ranking}
\end{center}
\end{table*}

\begin{table*}[!h]
\begin{center}
 \scalebox{0.70}{
  \begin{tabular}{ll|ll} \toprule
    \textbf{Rank} & \textbf{Headline} & \textbf{Rank} & \textbf{Headline}\\ \cmidrule{1-4}
     1 & Dow Drops 176; Nasdaq Tumbles 179  & 14 & Stocks, Dollar Lower on Strong Economic Report ...\\
     2 & U.S. stocks drop as bond market signals slowdown; Dow ... & 15 & Dow Down 60.50; Nasdaq Off 8.78 \\
     3 & U.S. stocks drop on profit-taking, poor Time Warner ...  & 16 & Stocks dip as traders await Fed meeting details\\
     4 & Dollar Lower, Stocks Fall in Early Trading & 17 &Dow Drops 17; Nasdaq Down Fraction\\
     5 & Stocks fall despite manufacturing pickup ...  &18 & Dollar Weaker, Stocks Fall ...\\
     6 & Nasdaq Falls 95; Dow Up 8 & 19 & Dollar, Stocks Traded Lower Eds ...\\
     7 & Stocks Fall, Dollar Traded Higher ... & 20 & Stocks Fall Back, Dollar Lower ...\\
     8 & Stocks fall in early trading &21 & Stocks fall on concerns over Wall Street and local ec...\\
     9 & U.S. stocks turn lower as investors take profits from ... &22 & Dollar, Stocks Lower in Early Tokyo Trading ...\\
     10 & Dow closes below 10,000; Nasdaq at lowest level ... &23 & Nasdaq Ends Down 147; Dow Up 25\\
     11 & Stocks lower on Wall Street amid mixed global picture ... &24 & U.S. stocks end mostly lower after GDP report ...\\
     12 & Financial shares fall on delay, restatement of results &25 & Dow Up 70.20; Nasdaq Falls 71.28 \\
     13 & Stocks Plunge on Profit-Taking, Dollar Inches Higher ... &26 & London Shares Lower ...\\
     \bottomrule 
  \end{tabular}
  }
 \caption{Top results for S\&P500}
  \label{tab:global_ranking_SP500}
\end{center}
\end{table*}

\begin{table}[!h]
\begin{center}
 \scalebox{0.70}{
  \begin{tabular}{ll} \toprule
    \textbf{Rank} & \textbf{Headline} \\ \cmidrule{1-2}
1 & Latam stocks lower on slowdown concerns\\ %; manufacturing reading boosts confidence\\

2 & Stocks end lower amid worries after House OKs plan\\% slide; Nike profit gives market a lift\\

3& Latam stocks plunge on slowdown concerns\\%, McCain meets with Gordon Brown in London\\

4&World markets drop on worries of US-led slowdown \\

5&French economy enters recession\\

6&US economy sheds most jobs since 2003\\

7&Manhattan apartment sales drop further\\%ernment struggling with rising threats\\

8&India ' s key stock index drops 4 percent\\

9&Japan stocks slide on worries about US economy\\%unveils  %probe into 'false rumors'\\

10&Hon Kong stocks drop on US worries\\ %when he returns\\

11&Credit markets still tight after bailout approval \\%after report shows spike in jobless claims\\

12&US cuts off family planning group in Africa\\

13&Employers cut 159,000 jobs, most in over 5 years\\

14&Russian shares fall sharply\\

15&US Congress OKs bailout bill and Bush signs its\\

%15&Commodities prices plunge on stronger dollar, fund selling\\

%16&Oil prices drop, near mid-US\$102 after US report shows oil, gasoline demand softening\\

%17&White House press secretary fumbles argument about Bush, says people required not to like him\\

%18&JPMorgan Chase makes it difficult for third parties to make offer for Bear Stearns\\

%19&EUROPE NEWS AT 1200GMT\\

%20&Barnsley able to field Steele in FA Cup semifinal after signing him on loan for rest of season\\
     \bottomrule 
  \end{tabular}
  }
 \caption{Top 15/219 -- S\&P500 -- October 3, 2008}
  \label{tab:daily-rank}
\end{center}
\end{table}

{\setlength{\parindent}{0cm}{
\textbf{Scoring news headlines.} Now we need to understand if the attention weights are consistent with results in Table~\ref{stock_prediction}; we expect 'business' headlines to have a relatively higher score. We trained the network on the entire set of news and used the attention layer to score 271,520 headlines across all categories in the test set. We selected the model with the minimum loss, patience of two epochs.
Table~\ref{tab:news-ranking} shows the results for the top 10, 100, 1,000, and 2,500 headlines. It shows the fraction of business headlines up to that rank, and the increase compared to the fraction of business news in the whole set. For the indices with higher accuracy in the previous experiment (S\&P500, Dow Jones, and Russell 1000), the scores significantly skew the distribution at the top ranks towards business news by between 534.65\%--471.09\% at Rank 10 and between 419.31\%--84.14\% at Rank 2500. For Nasdaq, with a previously lower performance, the scores do not seem to provide a clear pattern, indicating that the stock prediction performance might reflect the quality of scores. 
}}

\subsection{Anecdotal data}

Table~\ref{tab:global_ranking_SP500} shows the top 26 headlines over the whole timespan, ranked using the unnormalized attention layer weights of the model trained for S\&P500 with all news categories. The examples show that the model scores market-relevant headlines highly. We see mostly headlines reflecting general market trends. Results for an iconic date, October 3, 2008, in which the US House passed the 2008 bailout show the same trend (Table \ref{tab:daily-rank}). As for a single day, specific news about stock movements are not many, top-ranking has space for other relevant economic or political events.

\subsection{Manual evaluation}

We labeled the top 200 test set headlines for each index plus 200 uniformly sampled. Two annotators classified them as relevant or non-relevant. In total, there were only 19 (1.9\%) discrepancies that were resolved via mutual agreement. Table \ref{manual_evaluation_results} shows the relevance results and the high inter-annotator agreement. As before, financially relevant news score higher for the best performing indices compared to Nasdaq and the uniform sample. 

\begin{table} [!h]
\begin{center}
 \scalebox{0.70}{
  \begin{tabular}{lcccc} \toprule
    \textbf{Stock Index} & \textbf{Relevant} & \textbf{Not Relevant} & \textbf{Cohen's kappa}\\ \cmidrule{1-4}
     S\&P500  & \textbf{100}\% & 0\% & 1 \\
     Dow Jones  & \textbf{89\%} & 11\% & 0.88\\
     Russell 1000  & \textbf{87\%} & 13\%  & 0.94\\
     Nasdaq  & 25.5\% & \textbf{73.5}\%  & 0.95\\
     \hline
     Uniform  & 19\% & \textbf{81\%}  & 0.90\\
     \bottomrule 
  \end{tabular}
  }
 \caption{Manual evaluation on top 200 headlines}
  \label{manual_evaluation_results}
\end{center}
\end{table}
\vspace{-0.5cm}

\section{Conclusion and future work}

We presented an exploratory analysis to rank financially relevant events without manually labeled data. We showed that when a simple neural network is able to extract informative signals from news, the attention layer was able to score higher the most relevant news. Future work needs to focus on a more fine-grained analysis of the data and understanding the stability of the scores.

\section*{Acknowledgments}

We gratefully acknowledge the support of NVIDIA Corporation with the donation of the Titan V GPU used for this research.

\bibliography{anthology,eacl2021}
\bibliographystyle{acl_natbib}

\end{document}